\documentclass[lettersize,journal]{IEEEtran}
\usepackage{amsmath,amsfonts}
\usepackage{algorithmic}
\usepackage{algorithm}
\usepackage{array}
\usepackage[caption=false,font=normalsize,labelfont=sf,textfont=sf]{subfig}
\usepackage{subcaption}
\usepackage{textcomp}
\usepackage{stfloats}
\usepackage{url}
\usepackage{verbatim}
\usepackage{graphicx}
\usepackage{cite}
\usepackage{comment}
\usepackage{pifont}
\newcommand{\cmark}{\ding{51}} 
\newcommand{\xmark}{\ding{55}} 

\begin{document}
\title{Data-Efficient Approach to Humanoid Control via Fine-Tuning a Pre-Trained GPT on Action Data}

\author{Siddharth Padmanabhan$^{1}$, Kazuki Miyazawa$^{1}$, Takato Horii$^{1}$, Takayuki Nagai$^{1,2}$

\thanks{$^{1}$Graduate School of Engineering Science,
        Osaka University, 1-3 Machikaneyama-cho, Toyonaka-shi, Osaka 560-8531, Japan}
\thanks{$^{2}$AIX, The University of Electro-Communications, 1-5-1 Chofugaoka, Chofu-shi, Tokyo 182-8585, Japan}%
}
\markboth{Journal of \LaTeX\ Class Files,~Vol.~14, No.~8, August~2021}%
{Shell \MakeLowercase{\textit{et al.}}: A Sample Article Using IEEEtran.cls for IEEE Journals}


\maketitle

\begin{abstract}
There are several challenges in developing a model for multi-tasking humanoid control. Reinforcement learning and imitation learning approaches are quite popular in this domain. However, there is a trade-off between the two. Reinforcement learning is not the best option for training a humanoid to perform multiple behaviors due to training time and model size, and imitation learning using kinematics data alone is not appropriate to realize the actual physics of the motion. Training models to perform multiple complex tasks take long training time due to high DoF and complexities of the movements. Although training models offline would be beneficial, another issue is the size of the dataset, usually being quite large to encapsulate multiple movements. There are few implementations of transformer-based models to control humanoid characters and predict their motion based on a large dataset of recorded/reference motion. 
In this paper, we train a GPT on a large dataset of noisy expert policy rollout observations from a humanoid motion dataset as a pre-trained model and fine tune that model on a smaller dataset of noisy expert policy rollout observations and actions to autoregressively generate physically plausible motion trajectories. We show that it is possible to train a GPT-based foundation model on a smaller dataset in shorter training time to control a humanoid in a realistic physics environment to perform human-like movements.
\end{abstract}


\section{Introduction}
\IEEEPARstart{H}{umanoid} whole body control is becoming an increasingly interesting domain for enabling humanoids to be platforms for testing powerful deep learning models such as transformers for whole body control. However, for any character or robot that operates in continuous domain, reinforcement learning is time consuming and difficult for multiple reasons; designing reward statement is usually task-specific, character/robot specific and/or possibly simulator specific, and the online interaction between the training model and the simulator consumes time. Imitation learning has come a long way, and many papers have shown the efficacy of using motion capture data as reference data to train humanoids to learn natural human-like movements\cite{drrevisited, learningagile, tang2024humanmimic}. However, imitation learning by using kinematic data alone cannot be used to implement a model to control a humanoid in a realistic physics environment, as it serves to only help the model imitate the desired behavior and not necessarily capture the realism of the dynamics as well unless additional feedback like reward signals are used. This impact of the trade-off between reinforcement learning and imitation learning can be minimized by employing both strategies to encourage the model to imitate desired behaviors without having to design reward functions and at the same time follow the rules of physics either in simulators or in the real world.
It is particularly difficult to generalize motion for humanoids on a variety of tasks, either due to the complex nature of the skeleton structure and/or the task. Expressing a single controller that can perform multiple movements is holy grail for whole-body control of humanoids.

On the other hand, research is progressing on foundation models that can be adapted to multiple tasks \cite{bommasani2021opportunities}. For instance, in the field of natural language processing, a single model can perform multiple tasks by self-supervised pre-training on a large amount of textual data \cite{radford2019language}. Similarly, in humanoid control, if we can create a foundation model that can adapt to multiple tasks, we can improve humanoid adaptability.

\begin{figure} 
\centering
    \includegraphics[scale=0.4]{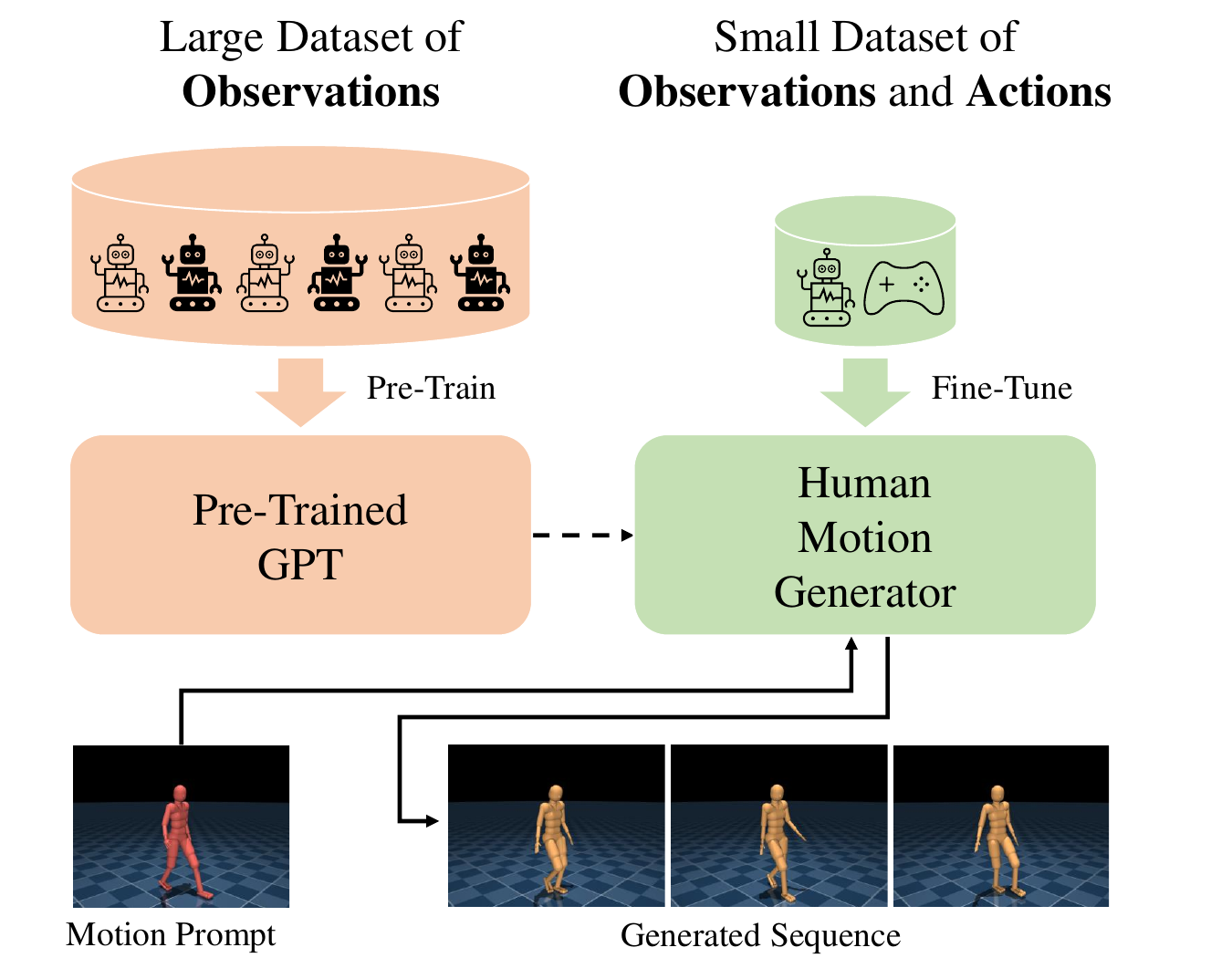}
    \caption{Proposed approach to training a GPT policy to autoregressively generate physically plausible motion in simulation by pre-training on a large dataset containing only observations and fine tuning on a small dataset containing observations and actions}
    \label{fig:HMG}
\end{figure}

In this study, we explore the potential of using a pre-trained motion foundation model, originally trained on non-physics data, and fine-tuning it on a smaller physics-based dataset. This approach allows us to avoid the extensive training time typically required for large multi-task models in an online simulator environment. Instead, we utilize a substantial dataset comprised of observations and actions derived from policy rollouts across multiple tasks. 
This facilitates data-efficient imitation learning. 

We hypothesize that by using a GPT-based motion pre-trained model, we would only have to fine tune this model on a smaller dataset, which we refer to as a Human Motion Generator (HMG), to plan physically plausible trajectories for humanoid motion, therefore significantly reducing the training time and dataset size (Fig. \ref{fig:HMG}).

In this paper we:
\begin{enumerate}
    \item demonstrate data efficient learning via training a GPT based on a large dataset without physics data and then fine tune on physics data.
    \item perform a comprehensive evaluation between our proposed model HMG and models trained from scratch by comparing the performances based on motion prediction metrics, trajectory generation lengths, empirical analysis of humanoid motion.
\end{enumerate}
We show from our experiments that our proposed model can generate more accurate and longer motion trajectories of higher quality than the state of the art. These promising results serve as a preliminary step towards developing the first humanoid motion foundation model for control that can be further fine-tuned for other downstream tasks.

\section{Related Work}
\subsection{Using Pre-trained Knowledge}
Generalization of human motion prediction is a difficult problem due to several reasons, such as the varying skeletal structure and idiosyncrasies in the human motion data. To tackle this issue, \cite{ImprovingHumanMotionPredictionThroughContinualLearning} trains a model through a curriculum and continual learning manner, such that a model can be first trained on a diverse dataset to be robust and then fine tuned to predict motion of new subjects.
\cite{zhu2023motionbert} proposes a two stage training, a pre-training stage where 3-D motion is derived from noisy partial 2-D observation, then a fine tuning stage where the model is fine-tuned to solve downstream tasks such as 3-D pose estimation, action recognition, and mesh recovery. \cite{radosavovic2023realworld} implements a similar version of a two-stage training, where initially the humanoid is trained to walk in a fully observed condition (having access to all sensory information and more from the humanoid) and then this policy is distilled to another policy trained in a partially observed condition. In this work, a GPT is pre-trained on observation data, resulting in a generalized representation of humanoid kinematics over a variety of of behaviors, and later fine tuned on action data that can control the humanoid in a physics simulator.

\subsection{Transformer based models for humanoid/robot control}
Transformers have been shown to be powerful to generate human motion\cite{radosavovic2023realworld, trajectorytransformer}. \cite{motiontransformer} uses the idea of dual attention to capture spatial and temporal dependencies of known data without relying on hidden states in RNNs or temporal encodings like Discrete Cosine Transformation. This model effectively generates poses that are temporally coherent. MotionGPT fuses language and motion to enhance performance of motion-related tasks such as text-to-motion, motion generation, and motion in-betweening \cite{motiongpt}. \cite{wagener2023mocapact} also demonstrates GPT's capability to generate motion in a physics engine after taking in one second long motion prompts. In this paper, we train a GPT policy to control a humanoid, similar to \cite{wagener2023mocapact}, but by employing a more data-efficient training strategy.

\subsection{Prediction Models for Humanoid}
Many papers have focused on optimizing human motion prediction by using various deep learning techniques\cite{worldmodels, dreamerv2, dreamerv3}. Instead of using memory-based networks like RNN or Transformers, a simple feed-forward network with fully connected layers, layer normalization,
and transpose operations, can be used to generate human motion using spatial and temporal information \cite{Back2MLP}. Similarly considering the idea that human motion prediction depends on spatial and temporal information, by using a scene image,
the past body poses of human, and the past 2-D locations as contexts, future 3-D poses and 3-D locations can be predicted \cite{LongTermHumanMotionPredictionwithSceneContext}. FrankMocap \cite{FrankMocap} proposes a modular approach where regression is performed for face, hands, and body individually and then integrated later to produce a whole body pose output. \cite{dlow} proposes DLow, a sampling strategy to obtain diverse set of samples from a trained generative model. This sampling strategy serves to tackle two problems; lack of diversity and inability to cover minor nodes in the data distribution. Here, we use GPT for motion generation since it is powerful for predicting long sequences of structured data given a small context.

\subsection{Datasets}
There are many popular datasets for human motion capture data available to the public\cite{h36m_pami, zhu2023h3wb, BABEL, AMASS}, but not many have simulation data consisting of control outputs coupled with observational data. In this paper we use both the large and small versions of the MoCapAct dataset\cite{wagener2023mocapact}, to train and evaluate our proposed model HMG and other GPTs for comparative evaluation. MoCapAct dataset is a dataset available to the public that consists of several rollouts of states and actions of the MuJoCo ball-joint humanoid played in the MuJoCo simulator. There are two versions of this dataset, one is around 500 GB in size and the other is 60 GB in size. In \cite{wagener2023mocapact}, a GPT  policy was trained from scratch whereas we show a more data-efficient approach to train a GPT via pre-training and fine-tuning, while also obtaining similar, if not, better motion generation capability.

\section{Methodology}
Pre-training a model and then fine-tuning that model on downstream tasks is a common practice adopted in NLP \cite{gpt3, BERT}, and in human motion prediction for human-robot interaction.
The architecture, datasets, and hyperparameters used are the same as those used in \cite{wagener2023mocapact}. 

\begin{figure*} 
\centering
    \includegraphics[scale=0.41]{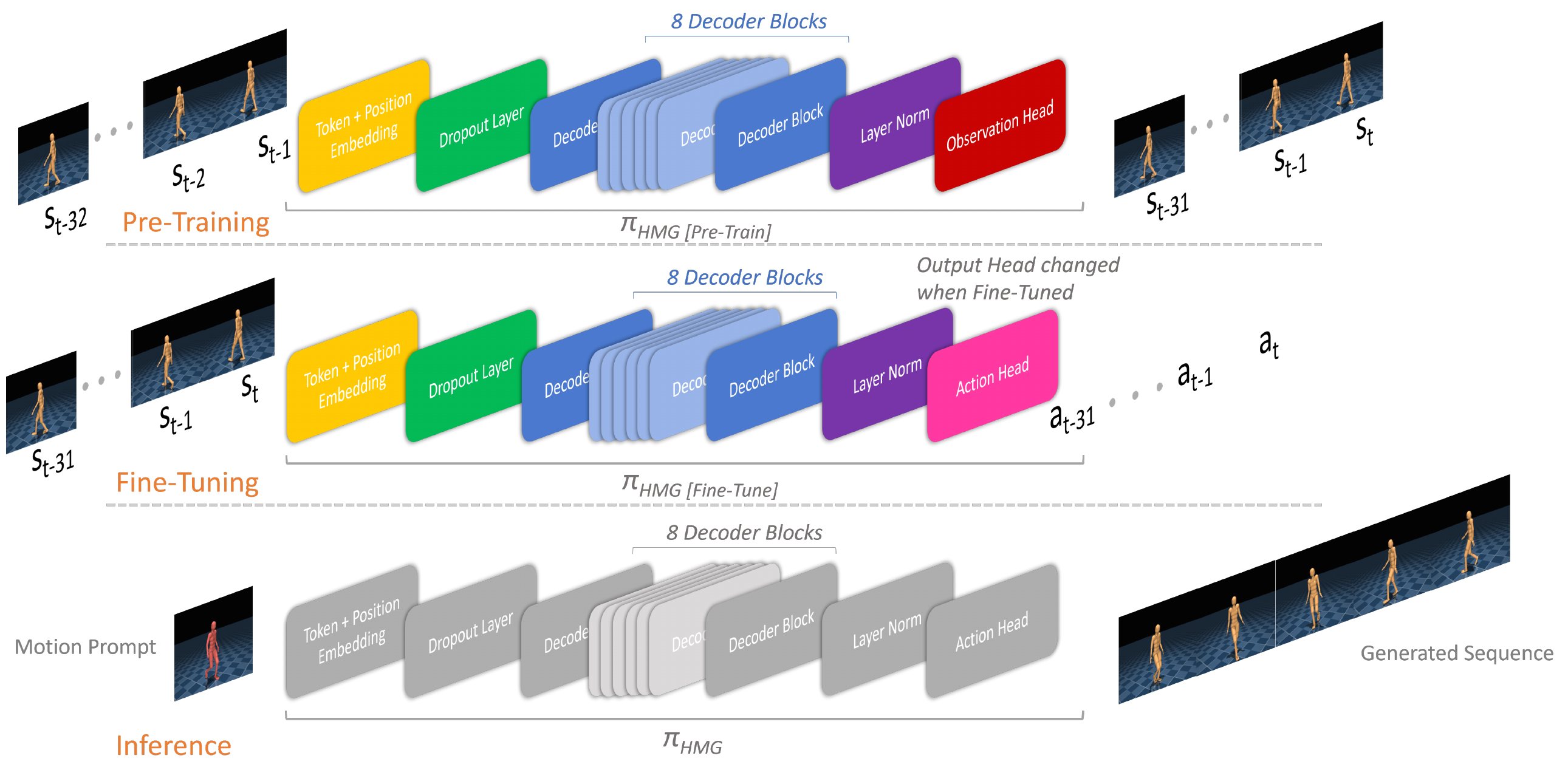}
    \caption{Detailed Proposed Approach of HMG: On top is the pre-training phase where a GPT is trained on a large observation dataset consisting of only observations; in the middle is the fine tuning phase, where the same GPT weights are used except the observation head which is replaced with an untrained action head to output actions, and the pre-trained model is fine-tuned on a small dataset consisting of both observations and actions; the bottom shows the inference of the resulting HMG. The GPT weights in gray depict that they are frozen, and the GPT weights in other colors denote that they are trainable.}
    \label{fig:HMFM}
\end{figure*}

The training methodology implemented and architecture are shown in Fig. \ref{fig:HMFM}. The first step is to train the humanoid motion foundation model. This serves as a pre-training phase. The motion foundation model is a minGPT \cite{minGPT} with model size 57M parameters. The motion foundation model is trained only on the observations taken from the large version of the MoCapAct dataset, i.e, the input and the output were observations. This dataset contains 100 rollouts of each motion behavior.

After the pre-training phase, this foundation model is fine-tuned by loading the previously obtained weights, and replacing the final linear feed-forward layer with a new learnable layer suited to output actions. Using both observations and actions taken from the small version of the MoCapAct dataset the foundation model is fine-tuned to realize the physics of the behaviors. This smaller dataset contains 10 rollouts of each motion behavior.

Later in the Experiment Section, we show in our evaluations of fine-tuning our pre-trained model on different dataset sizes and show that our proposed model surpasses models trained from scratch quicker with significantly smaller dataset size. A learning rate of $3*10^{-7}$, which is lower than that used during pre-training, was used in fine tuning.

The weights of the pre-trained model were updated against a cross-entropy loss during training and an MSE loss during validation. The losses have a minor change during the fine-tuning process where the losses are calculated between actions but not observations. The cross entropy loss is defined below
\begin{figure}[h]
    \centering
        \centering
        \includegraphics[scale=0.28]{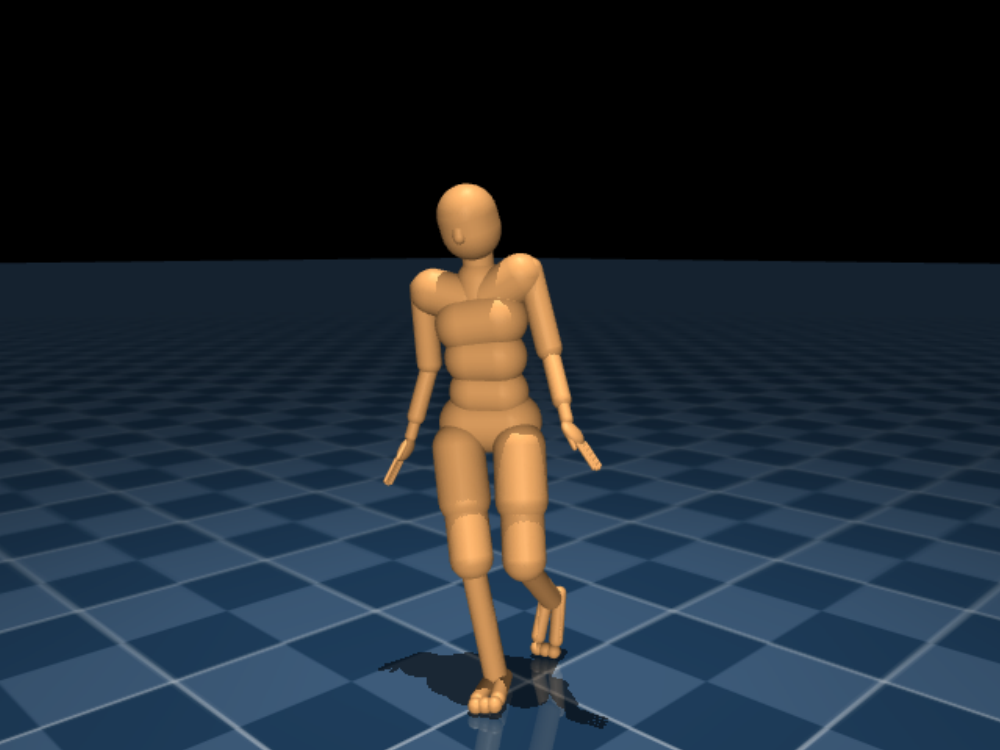}
        \caption{MuJoCo humanoid from dm\_control package in MuJoCo simulator}
        \label{fig:mujocohumanoid}
\end{figure}
\begin{equation}
\label{CEL}
    L_{ce} = \sum_{i=1}^{N}\hat\tau_{i} \log({\tau}_{i}),
\end{equation} where $\hat\tau_{i}$ can be either generated observation $\hat s_{t}$ or generated action $\hat a_{t}$ depending on whether the model is trained in pre-training stage or fine-tuning stage, and ${\tau}_{i}$ denotes either real observations $s_{t}$ or actions $a_{t}$. $N$ is the output size, therefore if $\tau_{i}$ is action then $N$ is the number of actions and if $\tau_{i}$ is observation then $N$ is the number of observations. The observable quantities used in training the pre-trained model and HMG is given in Table \ref{tab:observables}. The mode of control is positional control and the output actions are joint positions. Just like in \cite{wagener2023mocapact}, the simulator used in this work to test our motion prediction policy is MuJoCo\cite{todorov2012mujoco} and the humanoid used in the experiments is a standard MuJoCo humanoid from the dm\_control package that has 56 DoF (Fig. \ref{fig:mujocohumanoid}). 

\begin{table*}[h]
        \centering
        \caption{List of Observables and Actions taken from MoCapAct dataset}
        \label{tab:observables}
        \begin{tabular}{c|c}
        Observables & Description\\
        \hline
        Joint Pose & joint angles of each DoF in radians\\
        Velocimeter & root velocity in Cartesian XYZ directions\\
        Gyrometer & root orientation in Cartesian XYZ directions\\
        End Effector Pose & end effector orientation\\
        World Z Axis & direction of Cartesian Z axis\\
        Actuator Activation & boolean values on whether actuators are active or not\\
        Touch Sensors & contact forces between humanoid and ground\\
        Torque Sensors & joint torque of each DoF\\
        Body Height & root height from ground\\
        \end{tabular}
        \begin{tabular}{c|c}
        Actions & Description\\
        \hline
        Joint Pose & joint angles of each DoF in radians \\
        Control Mode & position control mode\\
        \end{tabular}
\end{table*}

\section{Experiment}
In our experiments, we train the pre-trained model on a large dataset and fine tune it on a small dataset for motion prediction downstream task, which we call it the HMG. We then compare and evaluate this HMG with MoCapAct-Small and MoCapAct-Large, where \textit{Small} and \textit{Large} denotes the size of the dataset the models are trained on.

The datasets used are the publicly available large and small versions of mocapact dataset. These datasets contains noisy observation and action data collected from expert policies. The pre-trained model is trained on a large dataset of only observations and then it is fine tuned on a smaller dataset consisting of both observations and actions. The time taken for pre-training is around 52 hours (2M steps), and the time taken for fine-tuning is around 12 hours (400K steps), thus the fine tuning phase takes almost a quarter of the time taken to train the motion foundation model.

We trained and tested three models: our proposed model HMG, MoCapAct-Large, and MoCapAct-Small, and performed a comprehensive comparative evaluation between them. The differences between these models in terms of training configuration is given in Table \ref{tab:models}. HMG is our proposed foundation model fine-tuned on a motion prediction downstream task. As previously stated, we chose two GPTs, namely MoCapAct-Large and MoCapAct-Small, trained on the large and small versions of the MoCapAct dataset respectively from scratch, to compare their performances with the performance of our model trained by using our proposed method on motion completion. Motion completion can be defined as the generation of motion based on a motion prompt input of a given length. The length of motion prompt used in this work is 32 steps that amounts to one second. The weights for MoCapAct-Large are publicly available and MoCapAct-Small was trained at our facility. We could see that although our proposed model and MoCapAct-Large possess similar capabilities, our model still slightly outperforms the latter.

The following subsections delineates the models' evaluations, where we see how well the models are capable of generating behaviors via prediction lengths, empirical differences in behaviors, and motion prediction metrics to evaluate motion quality, motion similarity to the ground truth, and motion diversity. Since the generated trajectory lengths by MoCapAct-Small were significantly lower than the other two models, we concluded that MoCapAct-Small performs poorly and did not include it in some of the evaluations.
\begin{table*}[t]
    \centering
    \caption{Models used for Evaluation}
    \label{tab:models}
    \begin{tabular}{c|c|c|c|c|c}
        & \multicolumn{2}{c}{Dataset} & \multicolumn{3}{|c}{Training}\\
        \hline
        Model & Large Dataset & Small Dataset & Pre-Trained & Fine-Tuned & Trained from scratch\\
        \hline
        HMG & \cmark (Pre-Training Only) & \cmark (Fine-Tuning Only) & \cmark (2M steps) & \cmark (400K steps) & \xmark\\
        MoCapAct-Large & \cmark & \xmark & \xmark & \xmark & \cmark (2M steps)\\
        MoCapAct-Small & \xmark & \cmark & \xmark & \xmark & \cmark (2M steps)\\
    \end{tabular}
\end{table*}

\subsection{Dataset Sizes}
\begin{figure}
\centering
    \includegraphics[scale=0.2]{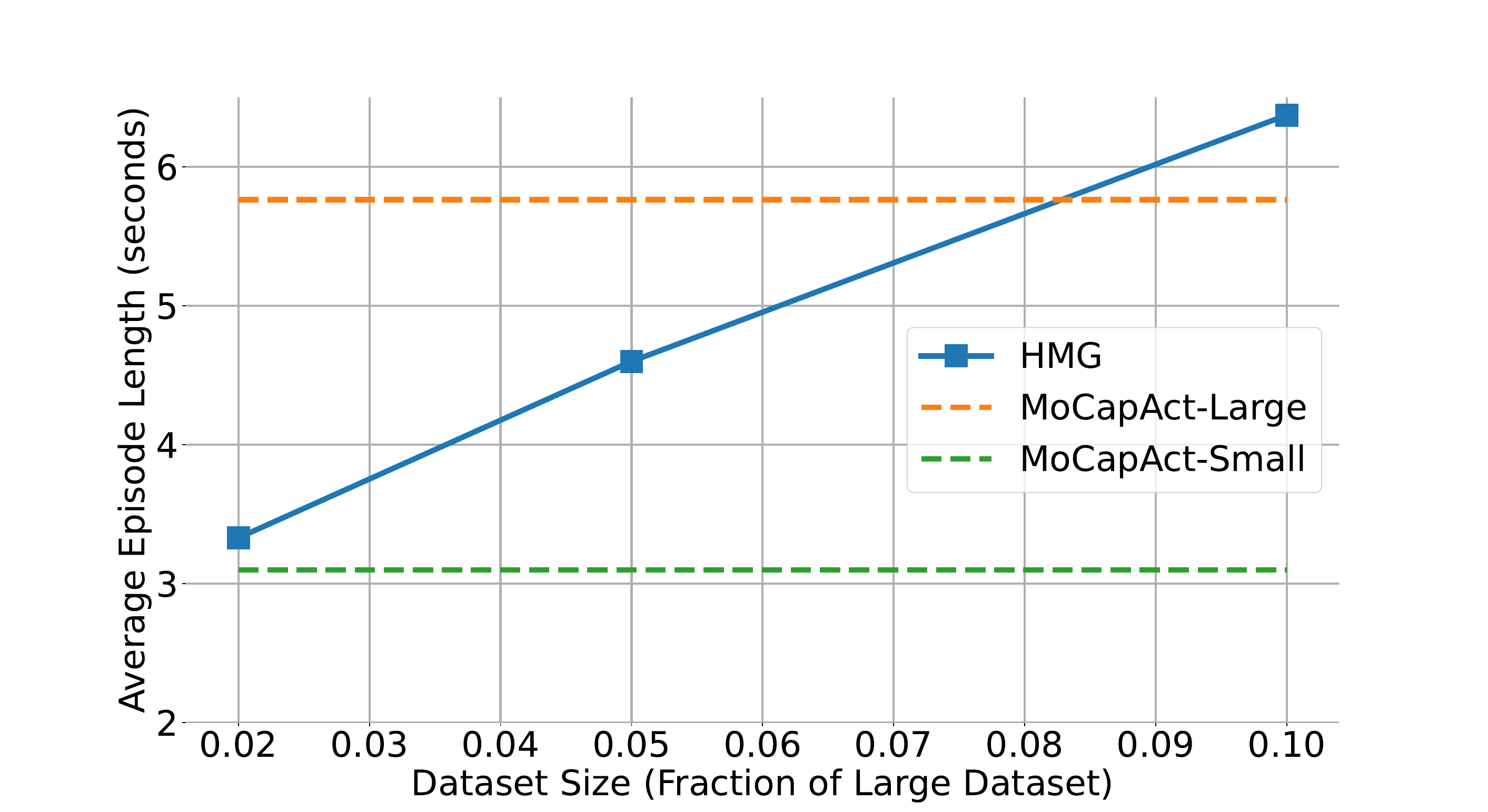}
    \caption{Performance based on Dataset Sizes}
    \label{fig:datasetsizes}
\end{figure}

Our proposed model was fine-tuned on datasets of half and quarter the size of the \textit{Small} dataset. The performances were compared based on episode lengths. We define an episode length as the total length of an episode for the humanoid to complete the task before episode termination (when it falls down) or the maximum episode length (set to about 15 seconds). We can see that the average generated episode lengths of the HMG trained on quarter and half the size of the \textit{Small} dataset are lower than that of MoCapAct-Large, which is trained on the \textit{Large} dataset that is 10 times the size of the \textit{Small} dataset, and higher than that of MoCapAct-Small. From Fig. \ref{fig:datasetsizes}, we can see that HMG quickly generates higher average prediction lengths on the validation datasets after being trained on the \textit{Small} dataset, compared to MoCapAct-Large which requires the \textit{Large} dataset to generate an average prediction length of 5.75 seconds. This shows that our proposed model is significantly data-efficient in learning motion prediction downstream task.

\subsection{Behavioral Differences}
\subsubsection{Quantitative Analysis}
Using the differences in episode lengths of motion predictions, we were able to find the exact behaviors generated by the models that outperformed the other models in terms of episode lengths. We considered the minimum difference in episode lengths of generated behaviors from the validation dataset (which consists of 63 behaviors), between models to be 6 seconds which we considered significant.

We observed that the number of episode lengths that were generated by HMG that were longer than that generated by MoCapAct-Large, were more than the number of episode lengths that were generated by MoCapAct-Large that were longer than that generated by HMG (Table \ref{tab:behavioraldifferences}). Also, the average of the differences in episode lengths were observed to be higher in HMG than in MoCapAct-Large. This shows that our proposed model has better capability in predicting longer trajectories that enables the humanoid to survive longer in the simulation environment.

\subsubsection{Qualitative Analysis}
\begin{figure*}
\centering
    \includegraphics[scale=0.36]{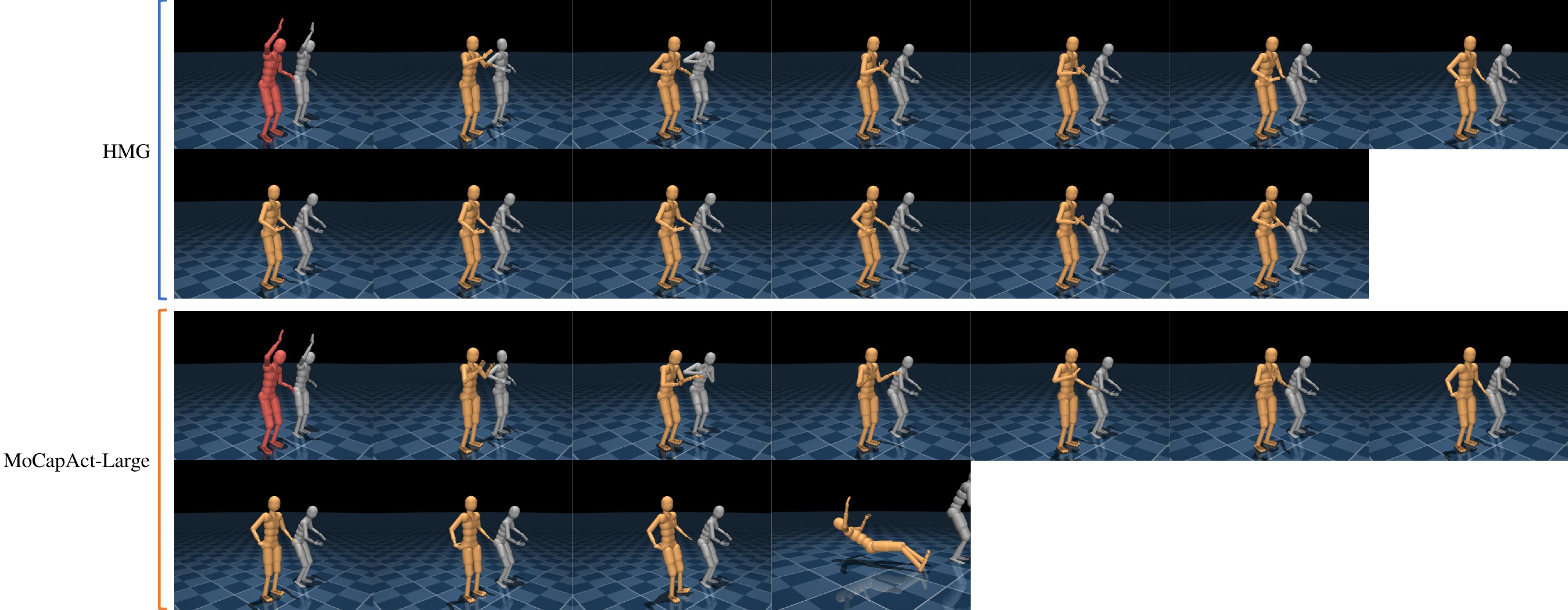}
    \caption{Gestures: Top - HMG's prediction does not match the ground truth but survives the entire duration of the episode., Bottom - MoCapAct-Large's motion generation also does not accurately predict the gestures, loses balance and falls.}
    \label{fig:gestures}

    \vspace{20pt}
    
    \includegraphics[scale=0.36]{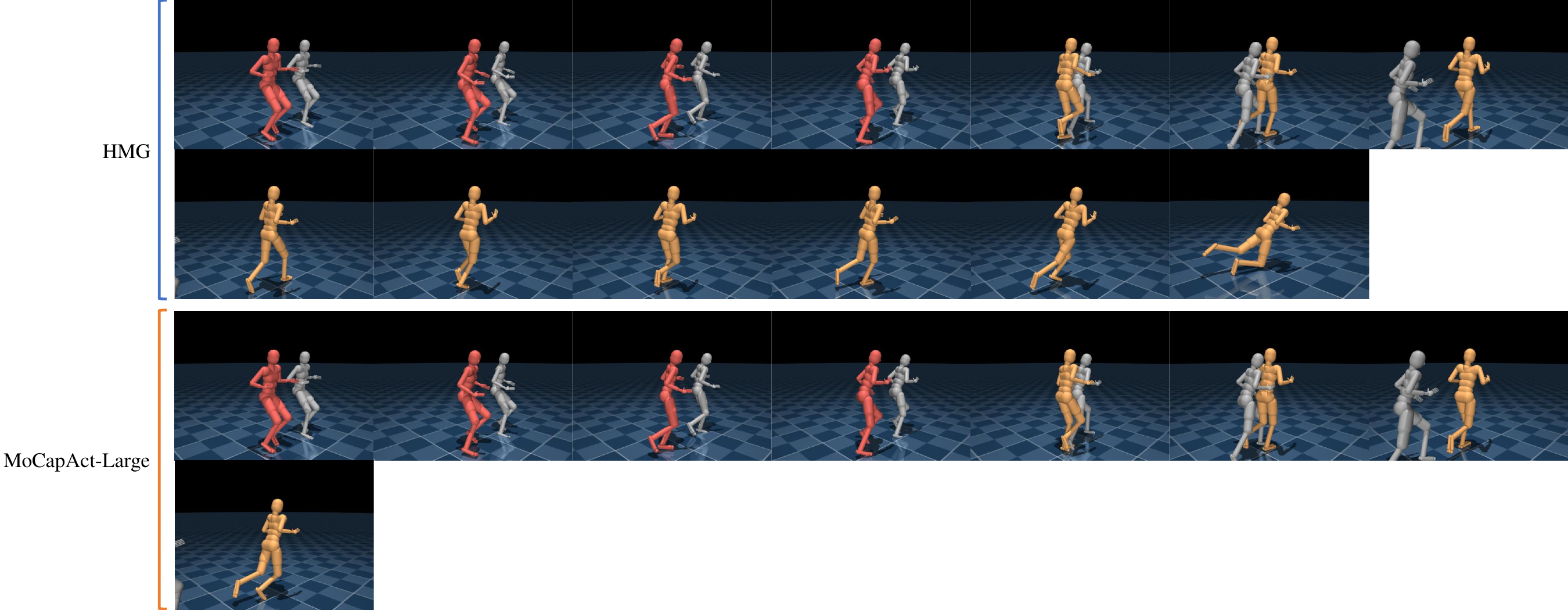}
    \caption{Running Towards the Left: Top - HMG survives a little longer than MoCapAct-Large but falls possibly due to the flight phase or fast changes in foot placement, Bottom - MoCapAct-Large falls almost immediately from the start of the motion prediction.}
    \label{fig:runningleft}

    \vspace{20pt}

    \includegraphics[scale=0.38]{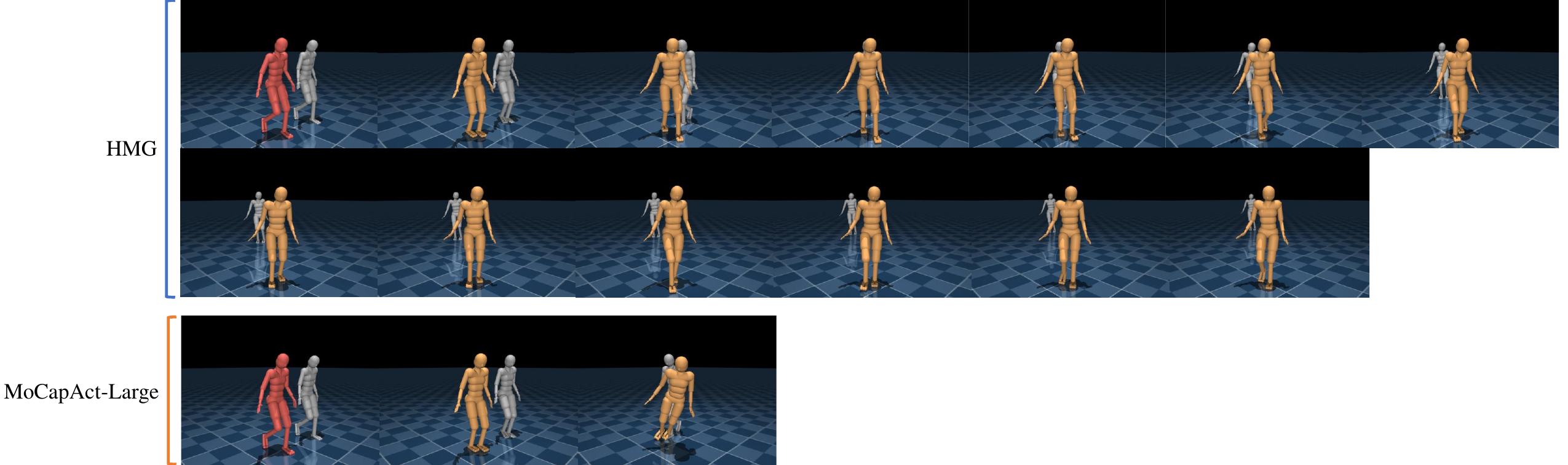}
    \caption{Walking Randomly: Top - HMG again maintains balance throughout the prediction while also succeeds in walking in different directions, Bottom - MoCapAct-Large fails to change direction and falls early.}
    \label{fig:walkingrandom}
    
\end{figure*}

We further looked into the exact behaviors that led to these differences in prediction lengths, to observe empirically how different the motion predictions were. Figs. \ref{fig:gestures} to \ref{fig:walkingrandom} show the comparison between the frames of the humanoid behaviors generated by HMG and MoCapAct-Large models. The gray humanoid depicts the reference motion, the other humanoid when in red denotes that the humanoid is given the motion prompt, and when in bronze denotes that the humanoid is under motion prediction. Fig. \ref{fig:gestures} shows a humanoid behavior that involves arm gestures while standing. Fig. \ref{fig:runningleft} shows a humanoid behavior that involves immediate change in direction and running. Fig. \ref{fig:walkingrandom} shows a simple locomotion behavior.

There were differences between the models' generations of moderate speed cyclic movements like walking, fast cyclic movements such as running, and non-cyclic movements as well. Walking behaviors patterns were observed to be similar, with the only difference being the generation length, with HMG coming out on top. Both the models suffer when predicting running behaviors and the humanoid falls quickly to the ground. We attribute this issue to the flight phase in running which might make it more difficult to predict the proper footstep planning. For non-cyclic behaviors like arm gestures or side-stepping, HMG generates movements that are not necessarily close to the ground truth, but helps it survive longer in the episode compared to MoCapAct-Large. This indicates that our proposed model has learnt different representation that abstract a better understanding of the relationship between observations and actions, that in turn helps in understanding the dynamics of the humanoid in the simulation environment to preserve the imitation and survival as much as possible.

\begin{table}[t]
    \centering
    \caption{Motion Generation Durability}
    \label{tab:behavioraldifferences}
    {
    \begin{tabular}{c|c|c}
          & No. of generated motions that& Avg. Episode \\
          & outlasted the other model & Length Differences \\ 
    Model & (from validation dataset)     & (sec) \\ 
    \hline
    HMG & \textbf{9 out of 63} & \textbf{8.630} \\
    MoCapAct-Large & 5 out of 63 & 8.082 \\
    \end{tabular}
    }
\end{table}

\subsection{Generated Trajectory Length}
In this subsection, we are going to cover the models' capability for generating long sequences given a motion prompt. HMG, MoCapAct-Large, and MoCapAct-Small were all made to generate predicted trajectories. The resulting episode lengths of generated trajectories of behavior clips from the training and validation datasets were recorded.The length of the motion prompts was 32 steps long, which constitutes one second of motion from the expert policy.

\begin{figure}
\centering
    \includegraphics[scale=0.2]{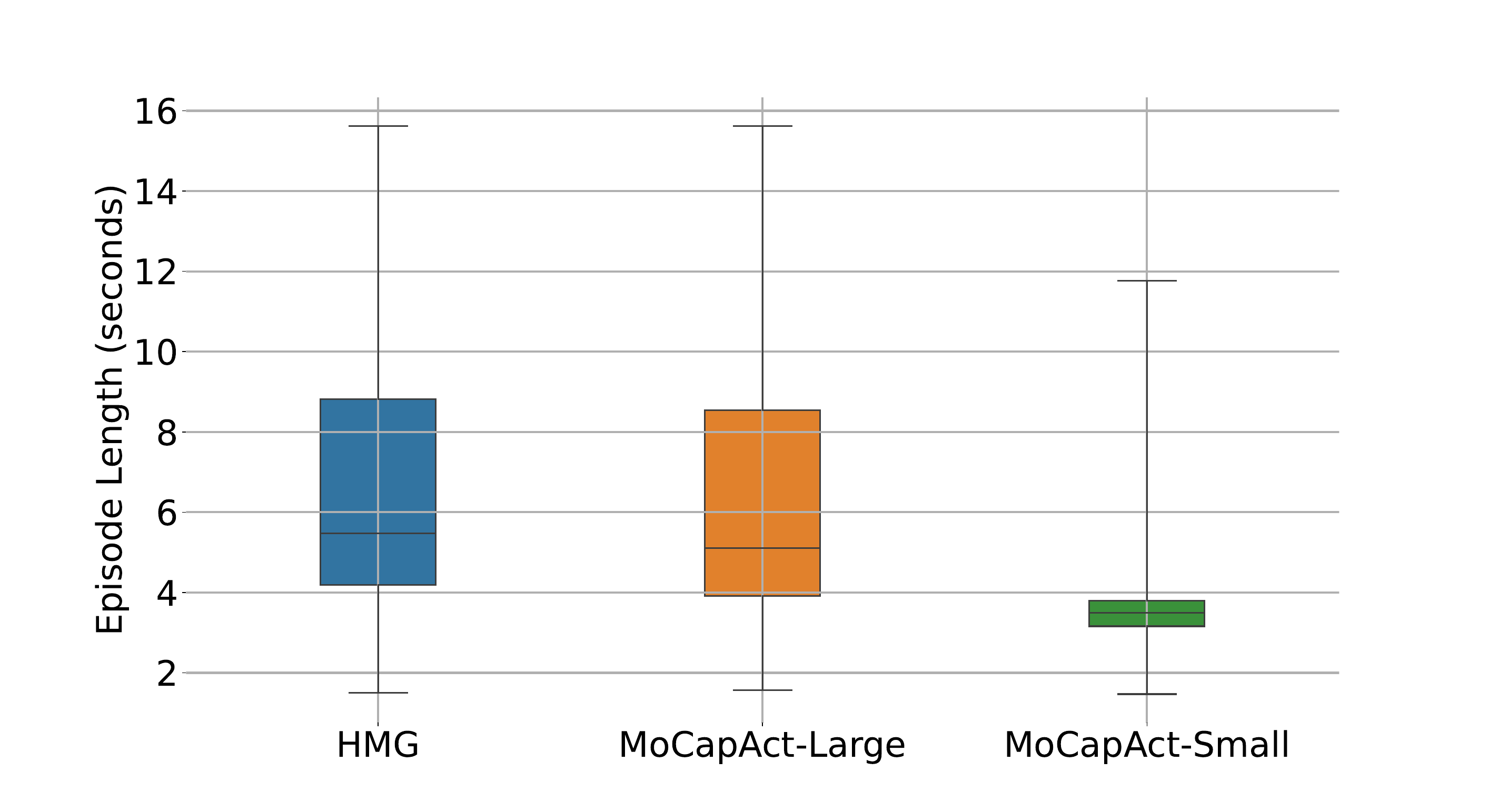}
    \caption{Comparison of Generated Episode Lengths on Training Dataset}
    \label{fig:Box Plots on Training Dataset}

    \vspace{20pt}
    
    \includegraphics[scale=0.2]{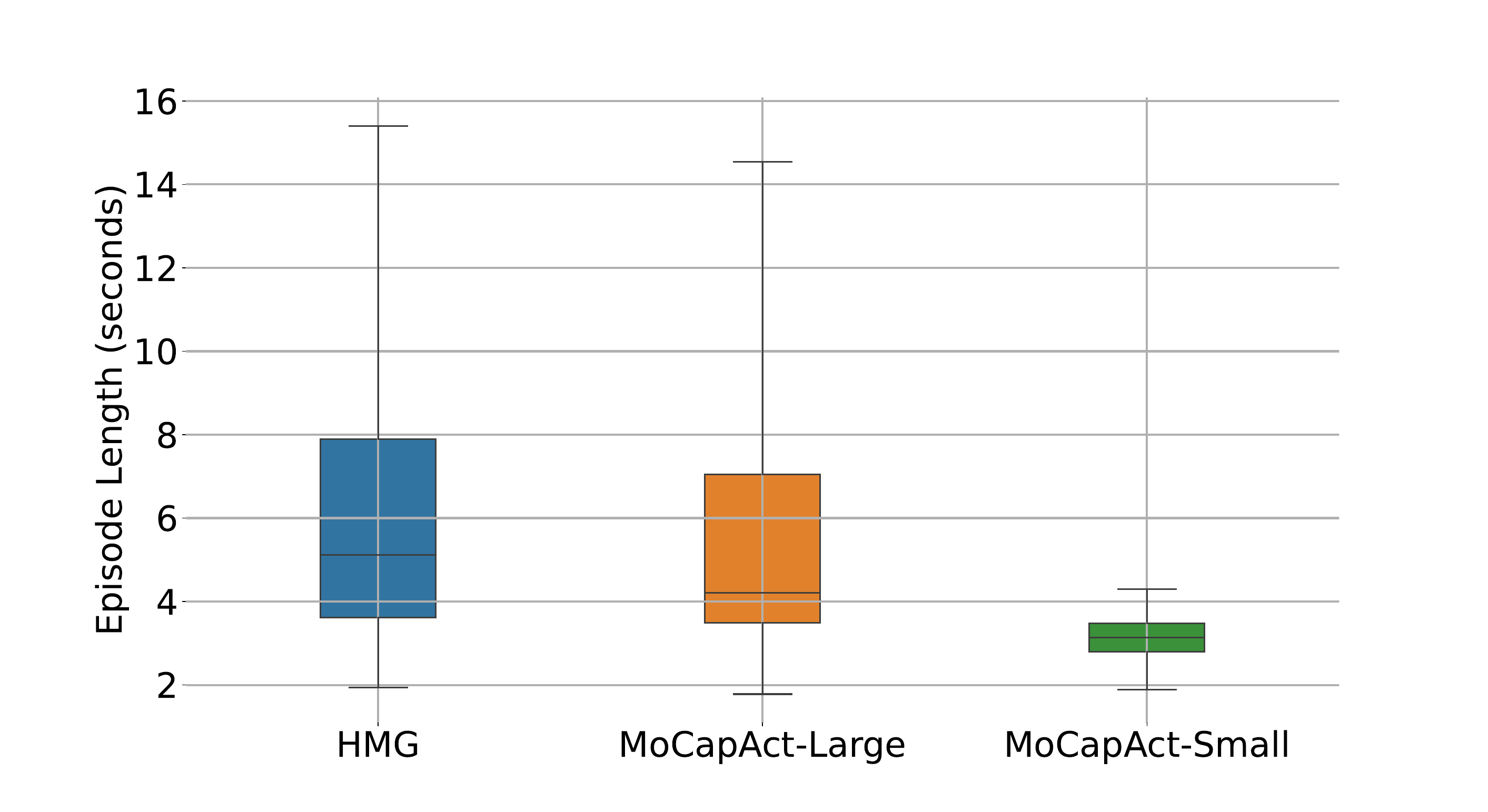}
    \caption{Comparison of Generated Episode Lengths on Validation Dataset}
    \label{fig:Box Plots on Validation Dataset}
\end{figure}

We can observe that the generated episode lengths of our proposed model are similar to those lengths generated by the MoCapAct-Large model, and overall slightly longer than MoCapAct-Large before episode termination. MoCapAct-Small, like the other models, performs well on the validation dataset compared to the training dataset. However, it fails to compete with the other models as it is unable to generate longer episode lengths. This shows that MoCapAct-Small lacks the ability to predict sequences greater than 5 seconds in length, indicating that training a GPT model on a small dataset for motion generation without pre-training performs poorly. We can confirm this from Figs. \ref{fig:Box Plots on Training Dataset} and \ref{fig:Box Plots on Validation Dataset} as well. From these box plots, it is easier to discern that HMG slightly outperforms MoCapAct-Large in terms of generating more behaviors and longer predictions, whereas MoCapAct-Small lack similar capabilities.

\subsection{Motion Prediction Metrics}
To further evaluate the motion prediction in terms of motion quality, motion imitation, and generation diversity, the standard metrics for motion prediction were chosen, namely; Frechet Inception Distance (FID), Average Displacement Error (ADE), Final Displacement Error (FDE), and Diversity (DIV). We chose these metrics as the standard based on previous works that used these metrics to evaluate kinematic motion prediction\cite{motiongpt, action2motion, heusel2018gans, motron}. We believe that these metrics are appropriate here even for dynamics motion prediction since the mode of control is positional (units are homogeneous between kinematic and dynamic motion prediction). First the metrics will be defined as below.

FID is one of the most important metrics to determine how well the generated motion quality is. The idea behind FID is to compare the distributions between the features from the motion feature extractor based on the real and generated trajectories. The common practice is to train another RNN as a motion feature extractor on the relevant data and then extract the features. Training another RNN in this case is redundant, since the GPT is trained on reconstructing and predicting the motion given the motion prompt, plus, the representation of the features taken from the middle of the GPT network is rich\cite{imagegpt}. Hence, the pre-trained model which was trained on observation data was used as the motion feature extractor to calculate FID and DIV. Equation (\ref{FID}) shows how to calculate FID
\begin{equation}
\label{FID}
\text{FID} = \|\mu_{r} - \mu_{g}\|^2 + \mathrm{Tr}(\Sigma_{r} + \Sigma_{g} - 2(\Sigma_{r}\Sigma_{g})^{1/2}),
\end{equation} where, $\mu_{r}$, $\mu_{g}$, $\Sigma_{r}$, $\Sigma_{g}$ are the means and co-variances of the features from the real and generated data respectively. $\mathrm{Tr}$ is the trace of the resulting matrix.

Understanding ADE is intuitively quite straightforward. It is essentially the average of the differences in joint poses across entire trajectories between the generated trajectories and the ground truth
\begin{equation}
\label{ADE}
\text{ADE} = \frac{1}{N \times T} \sum_{i=1}^{N} \sum_{t=1}^{T} \| \hat{\mathbf{j}}_{i,t} - \mathbf{j}_{i,t} \|,
\end{equation} where $\hat{\mathbf{j}}_{i,t}$ and $\mathbf{j}_{i,t}$ are the generated and real joint poses respectively at time step $t$ and trajectory $i$.

FDE is the difference in joint poses at the last step of the trajectory between the generated motion and ground truth
\begin{equation}
\label{FDE}
\text{FDE} = \frac{1}{N} \sum_{i=1}^{N} \| \hat{\mathbf{j}}_{i,T} - \mathbf{j}_{i,T} \|,
\end{equation} where $\hat{\mathbf{j}}_{i,T}$ and $\mathbf{j}_{i,T}$ are the generated and real joint poses respectively at the final time step of the trajectory $i$. ADE and FDE scores are in radians.

FID, ADE, and FDE help us understand the quality of the motion produced and how close the generated motion is to  the ground truth.

Another important metric is the generation diversity (DIV). This metric shows the variance in the generated motions across all behaviors the model is trained on. It is preferred if the variance is close to that of the real dataset. DIV is calculated by creating two sets of $n$ randomly sampled motions from each generated data and calculating the diversity between the two. Equation \ref{DIV} calculates the diversity score of a model
\begin{equation}
\label{DIV}
\text{DIV} = \frac{1}{n}\sum_{i=1}^{n} \mathbf{E_{d}}(\mathbf{v_i},\mathbf{v_{i}^{\prime}})
\end{equation} where $\mathbf{v}, \mathbf{v^{\prime}}$ are feature vectors from two sampled sets from the same generated models (expert policy, HMG, MoCapAct-Large) respectively across all behaviors, $n$ is the number of samples in the set, and $\mathbf{E_{d}}$ is the Euclidean distance. We set $n$ to be $200$. We require ADE and FDE to be low so that the motion similarity is high and we require FID to be low to ensure that the motion quality is as close to that of ground truth. From Table \ref{tab:motionpredictionmetricevaluation}, we can see that the model displays the desired scores except for DIV which deviates from the real data.
\begin{table}[t]
    \centering
    \caption{Motion Prediction Scores: $\downarrow$ indicates that lower score is desired and $\rightarrow$ indicates that score closer to \textbf{Real} (ground truth data) is desired.}
    \label{tab:motionpredictionmetricevaluation}
    {
    \begin{tabular}{c|c|c|c|c}
    Model & FID $\downarrow$ & ADE $\downarrow$ & FDE $\downarrow$ & DIV $\rightarrow$ \\
    \hline
    \textbf{Real} & 0.000 & N/A & N/A & 7.530 \\
    \hline
    HMG & \textbf{7.741} & \textbf{6.824} & \textbf{6.616} & 8.222 \\
    MoCapAct-Large & 8.415 & 6.866 & 6.910 & \textbf{8.151} \\
    \end{tabular}
    }
\end{table}

\section{Discussion}
In \cite{wagener2023mocapact}, box plots and histograms were used to evaluate their model's capability to generate motion.
However, since it is hard to tell how well the humanoid can actually imitate and complete the ground truth motion prompts accurately, the standard metrics FID, DIV, ADE, and FDE were chosen to evaluate the motion quality, imitation and diversity\cite{motiongpt, zhang2021joints, hmddistillation}.
From our evaluations, we could confirm that our proposed method of fine-tuning a pre-trained model is more data-efficient than training from scratch, and the imitation performance between both the methods are quite similar with our model slightly outperforming the other. However, we cannot control the generation of the motion by conditioning, for instance, if the model generates a motion given a prompt to stand, it may try to either continue to stand or walk. Another limitation is that it uses fully observable state space, whereas realistically we would prefer to only use partially observable variables. Other limitations include imperfect imitation and can probably perform better when fine tuned with reward feedback. We are curious to see if our foundation model can be trained and fine-tuned on simulation data obtained from other robots such as \cite{locomujoco}.

\section{Conclusion}
We proposed a data-efficient approach for motion prediction by pre-training a humanoid motion foundation model on observation data and fine-tuning it on both observation and action data on a motion prediction downstream task. We have evaluated our proposed method's performance with training from scratch based on several facets; training efficiency, motion prediction metrics, generation lengths, and empirical observations on various cyclic, non-cyclic, and fast paced behaviors. We conclude that our proposed method is more data-efficient than the conventional approach.

\section*{Acknowledgment}
This work was supported by the JST Moonshot R\&D Grant Number JPMJMS2011.

\bibliographystyle{IEEEtran}
\bibliography{main}

\end{document}